\documentclass[10pt,onecolumn, letterpaper]{article}

\usepackage{times}
\usepackage{epsfig}
\usepackage{graphicx}
\usepackage{amsmath}
\usepackage{amssymb}
\usepackage{booktabs}

\usepackage{multirow}
\usepackage{bbm}
\usepackage{algorithm}
\usepackage{algpseudocode}

\usepackage[pagebackref,breaklinks,colorlinks]{hyperref}

\usepackage[capitalize]{cleveref}
\crefname{section}{Sec.}{Secs.}
\Crefname{section}{Section}{Sections}
\Crefname{table}{Table}{Tables}
\crefname{table}{Tab.}{Tabs.}

\begin{document}

\title{PercentMatch: Percentile-based Dynamic Thresholding for Multi-Label Semi-Supervised Classification}


\author{
Junxiang Huang\thanks{Work done during an internship at Autodesk, Inc.}\\
Northeastern University\\
{\tt\small imjunxhuang@gmail.com}
\and
Alexander Huang\\
Autodesk, Inc.\\
{\tt\small alexander.huang@autodesk.com}
\and
Beatriz C. Guerra\\
Autodesk, Inc.\\
{\tt\small beatriz.guerra@autodesk.com}
\and
Yen-Yun Yu\\
Autodesk, Inc.\\
{\tt\small yen-yun.yu@autodesk.com}
}

\maketitle

\begin{abstract}
   While much of recent study in semi-supervised learning (SSL) has achieved strong performance on single-label classification problems, an equally important yet underexplored problem is how to leverage the advantage of unlabeled data in multi-label classification tasks. To extend the success of SSL to multi-label classification, we first analyze with illustrative examples to get some intuition about the extra challenges exist in multi-label classification. Based on the analysis, we then propose PercentMatch, a percentile-based threshold adjusting scheme, to dynamically alter the score thresholds of positive and negative pseudo-labels for each class during the training, as well as dynamic unlabeled loss weights that further reduces noise from early-stage unlabeled predictions. Without loss of simplicity, we achieve strong performance on Pascal VOC2007 and MS-COCO datasets when compared to recent SSL methods.
\end{abstract}



\section{Introduction}
\label{sec:intro}


Deep learning models have achieved superior performance for various computer vision applications, partially thanks to their ability to scale up with large amount of training data \cite{DBLP:journals/corr/abs-1712-00409, mahajan2018exploring, raffel2020exploring}. Even though it has been empirically confirmed that deep learning models gain stronger performance through supervised training with a larger dataset, the benefits of such performance improvement accompany a significant cost: generating high-quality annotations is an expensive and laborious process, especially when domain expertise is required \cite{gil2018classification, DBLP:conf/cvpr/WangPLLBS17, elezi2022not}. To alleviate the strong reliance on labeled data, semi-supervised learning (SSL) methods have been developed for the past decades to improve model performance by utilizing a tremendous amount of unlabeled data along with labeled data \cite{zhu2005semi, yang2021survey}. 

There are two popular paradigms that recent SSL methods have adopted: \textit{pseudo-labeling} which uses a model's class predictions as artificial labels to train successor models \cite{lee2013pseudo, Yalniz2019BillionscaleSL, gong2016multi, berthelot2019mixmatch}, and \textit{consistency learning} which enforces models to output consistent predictions on similar images \cite{Laine2017TemporalEF, Sajjadi2016RegularizationWS}. Some modern SSL methods have shown that the combinations of these two paradigms produce outstanding results for single-label classification tasks \cite{DBLP:conf/nips/XieDHL020, sohn2020fixmatch, zhang2021flexmatch}.

\begin{figure*}
\begin{center}
\includegraphics[width=.97\linewidth]{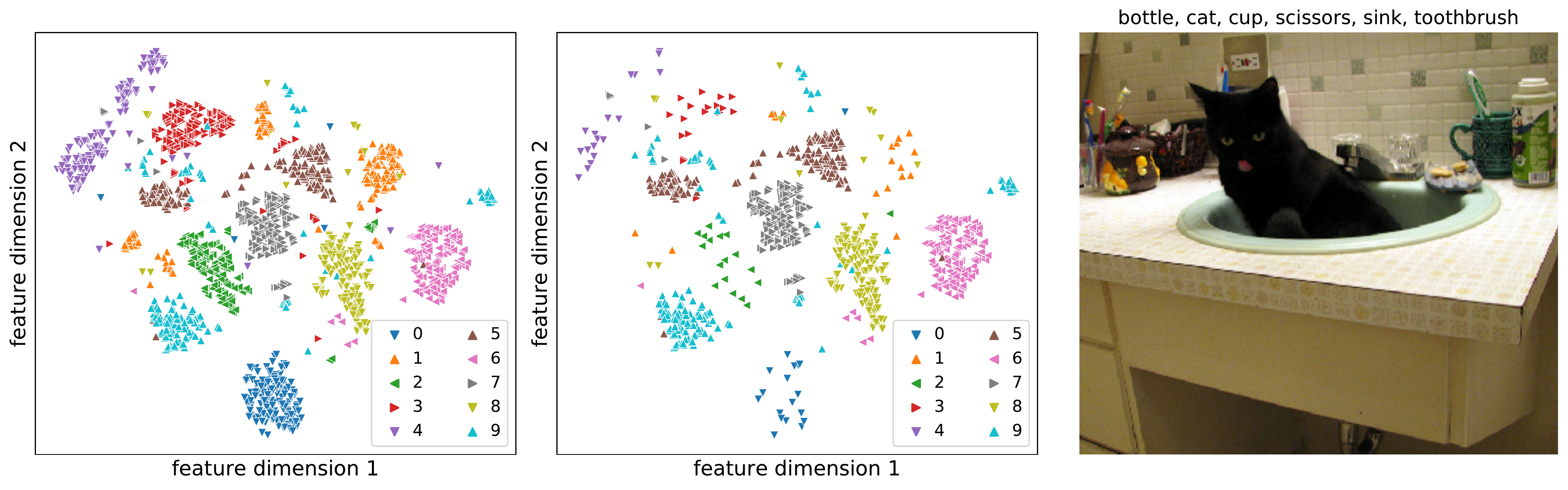}
\end{center}
	\caption{Semi-supervised methods for multi-label classification problems have to deal with extra challenges than that for single-label classification. (a) An example visualization of feature embedding of MNIST test set using t-SNE method \cite{van2008visualizing}. The 10 classes are perfectly balanced, and with a suitable encoder the samples with same label are well clustered. The two features are also valid for many single-label datasets. (b) We artificially create class imbalance in MNIST by randomly removing $90\%$ samples from classes \textit{0-4} and then visualize using the same method. (c) shows a sample image from MS-COCO (ID: 054918), in which the class ``scissors'' only occupies a negligibly small fraction of pixels.}
	\label{fig:embed}
\end{figure*}

Despite its application importance and inherent generality, how to extend the success of SSL to multi-label classification does not receive enough attention and investigation \cite{DBLP:conf/miccai/LiuTCBRC21,rizve2021in}. In order to bring all potentialities of SSL methods into full play, several extra challenges in multi-label classification need to be addressed: (1) Class presence become independent, which disables information achievement by method of exclusion. In single-label classification, even if a classifier learns terribly on one class, it is still able to correctly recognize positive samples of that class as long as it gives low scores to the other classes. In other words, given the number of classes $C$, method of exclusion increases the probability of getting correct pseudo-labels by discarding some unlikely choices from the total $C$ possible outcomes. But class independence in multi-label tasks turns the number of all possible outcomes into $2^C$ and invalidates this method. (2) A common assumption many SSL built on is that the classification decision surface should favor low-density areas of the marginal data distribution. However, if the distribution density variance between different classes is huge, which is common for multi-label datasets, an adaptive density thresholding is needed for this assumption. In Fig. \ref{fig:embed}(a), data points from MNIST \cite{lecun1998gradient} are visualized as an example of single-label dataset with good class balance, on which a decent decision boundary can be generated by avoiding the high-density cluster regions. However, when we artificially introduce class imbalance by down sampling half of the classes in Fig. \ref{fig:embed}(b), applying the same method based on vari could lead to decision boundaries. As multi-label datasets are typically more class-imbalanced than single-label ones, class-specific thresholds become more critical in SSL tasks. (3) Unlike single-label dataset, where the major part of a image usually represents the target class, multi-label targets may only occupy a small fraction of pixels in an image they appear, such as class ``scissors'' in Fig. \ref{fig:embed}(c). While the example image is an \textit{easy} SSL material for classes ``cat'' and ``sink'', it is at the same time \textit{hard} SSL material for some other classes. This suggests that in multi-label SSL scenario, we should evaluate the learning difficulty in the level of sample-label pair, which is different from the sample level evaluation in single-label curriculum learning \cite{DBLP:conf/iclr/WuDN21}.

In this study, taking these special challenges into considerations, we propose a percentile-based threshold adjusting scheme for multi-label semi-supervised classification. The main contributions of this research include: (1) a simple dynamic threshold framework to reduce the noise induced by incorrect pseudo-labels; (2) a natural methodology that can generalize prior SSL methods from single-label classification to multi-label classification, (3) empirical experiments showing that our method has outperformed the previous results on the Pascal VOC2007 \cite{Everingham10}, and new benchmark results on MS-COCO \cite{lin2014microsoft} dataset.

\section{Related Works}

In this section, we will briefly review some prior SSL methods, focusing on models with paradigms of \emph{pseudo-labeling} and \emph{consistency learning}.

The key idea of \textbf{pseudo-labeling} to use a pre-trained teacher model to infer on unlabeled data and generate a set of corresponding artificial pseudo-labels, which are then mixed with real labeled data to train a student model \cite{lee2013pseudo, Yalniz2019BillionscaleSL}. For single-label tasks, the pseudo-label of an unlabeled sample is selected as the class with the maximum prediction score. However, as incorrect pseudo-labels will confuse models and lower the final performance, the central problem of this paradigm is to select as few incorrect pseudo-labels as possible getting as many correct ones at the same time. A common solution is to choose a high score threshold and select only samples with a maximum score exceeding the threshold. Within this paradigm, based on the direction of changing this score threshold, curriculum learning \cite{DBLP:conf/aaai/Cascante-Bonilla21} and anti-curriculum learning \cite{hacohen2019power} are proposed: the former feeds only \textit{easy} pseudo-labels, i.e. samples with high scores, to models first, then progressively includes \textit{hard} pseudo-labels, i.e. samples with low scores. The latter revert the order of easy and hard pseudo-labels.


The methodology of \textbf{consistency learning} consists to perturb each sample with different augmentations and encourage the model to make similar predictions on the perturbed instances. One example is the Pi-model \cite{Sajjadi2016RegularizationWS}, which performs two parallel stochastic augmentations on each unlabeled data and generates two prediction vectors on both disturbed versions. A stability loss is introduced to penalizes different predictions for the same input sample by taking the mean square difference between two prediction vectors. Temporal ensembling \cite{Laine2017TemporalEF} is based on the same principle with a more general framework. In short, it generates one augmented version for each sample and keeps track of the moving average of predictions, then encourages models to output predictions that are similar to the moving average on the same samples. MixMatch \cite{berthelot2019mixmatch} extends the Pi-model by generating predictions on different versions of each sample obtained via parallel stochastic augmentations, and then encouraging the predictions to be close to the averaged scores. 

Some recent SSL methods have combined these two SSL paradigms by using weak and strong augmentations. UDA \cite{DBLP:conf/nips/XieDHL020} and FixMatch \cite{sohn2020fixmatch} perform a weak stochastic augmentation and a strong augmentation on unlabeled data, and introduce a fixed high score threshold to select only pseudo-labels whose confidence scores exceed the threshold. A loss term is leveraged to lead the strongly augmented predictions get closer to the corresponding selected pseudo-labels. This score-based thresholding is critical as it helps to retain high-quality pseudo-labels and reduce confirmation bias. Given the importance of thresholding, FlexMatch \cite{zhang2021flexmatch} argues that as the learning difficulty varies for different classes, a dynamical threshold for each class would increase the SSL efficiency. The key insight of FlexMatch is that by estimate the class-specific learning status, the score threshold can be adjusted for each class to improve the performance on hard classes. Besides, Dash \cite{DBLP:conf/icml/XuSYQLSLJ21} also adjusts the score thresholds by gradually increasing the universal threshold for all classes as the training progresses. FreeMatch \cite{DBLP:journals/corr/abs-2205-07246} goes further by have the thresholds adjusted in a self-adaptive manner. AdaMatch \cite{DBLP:conf/iclr/XingAZP20} provides a unified solution of SSL and domain adaptation by having thresholds based on labeled data and ignoring unlabeled data distribution. As previous method achieve strong performance on single-label datasets, they might fail to consider the specific challenges in multi-label classification tasks.


In addition to dynamic thresholding technique, another direction to improve pseudo-labeling efficiency is \textbf{negative label learning}. The previous mentioned methods typically use one positive threshold per class to select of discard all predictions on a sample. However, such an approach throws away those extremely-low-score predictions which indicate the absence of a class with high confidence. In Uncertainty-aware Pseudo-label Selection \cite{rizve2021in}, the authors introduce a negative threshold so that if the confidence score for a prediction is sufficiently low, it is considered as a negative example. They further enhance this procedure by introducing positive and negative uncertainty thresholds which ensures that the selected pseudo-label (whether positive or negative) has low enough uncertainty. This was shown to improve pseudo-label accuracy. In \cite{Wang2022SemiSupervisedSS}, the authors used an entropy threshold in combination with a rank-based threshold to select negative labels as those that have a low enough confidence score and have high entropy. In \cite{tokunaga_iwana_teramoto_yoshizawa_bise_2020}, the authors assign negative pseudo-labels based on a weakly supervised signal.

\section{PercentMatch}

\begin{figure*}
\begin{center}
\includegraphics[width=.97\linewidth]{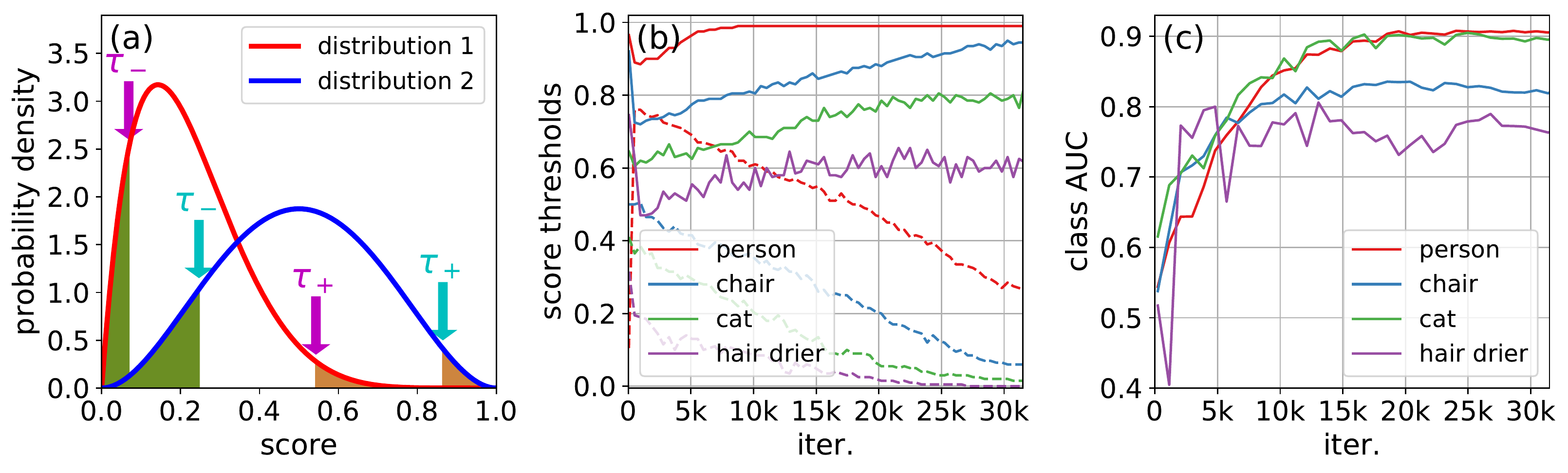}
\end{center}
   \caption{Illustration of percentile thresholds. (a) The red and blue curves are two different probability distributions of output scores, which can be different classes or same class at different training stages. Both the $\tau_-$ arrows correspond to $\kappa_-=0.1$, which means the area of each green shadow is $0.1$; and the $\tau_+$ arrows correspond to $\kappa_+=0.98$ for both distributions. (b) The trace of score thresholds for four sample classes on MS-COCO with $10\%$ labeled data and 224 input size. The score thresholds correspond to percentile thresholds of $\kappa_-=0.1$ (dashed lines) and $\kappa_+=0.98$ (solid lines). (c) shows the class ROC-AUC on the test set evaluated every $1$K iterations.}
\label{fig:thres}
\end{figure*}

Compared to FixMatch, PercentMatch has three main modifications addressing the special challenges in multi-label classification problems. Specifically, we introduce negative labels learning along with positive labels to select suitable sample-class pairs, replace fixed score thresholds with fixed percentile thresholds to accommodate different learning difficulties of classes, and introduce a learning-status-aware unlabeled loss weights to reduce noisy pseudo-labels.

In the following, we follow the notations used in \cite{rizve2021in}: Assume we have a labeled dataset of $N_L$ samples denoted by $\mathrm{D}_L=\big\{ (\mathbf{x}^{(i)}, \mathbf{y}^{(i)}) \big\}_{i=1}^{N_L}$, where $\mathbf{x}$ is the input, and $\mathbf{y}^{(i)}=[y_1^{(i)}, \dots, y_C^{(i)}] \in \{0, 1\}^C$ is the corresponding label with $C$ classes. For  $i$-th labeled sample, $y_c^{(i)}=1$ indicates the presence of class $c$ and $y_c^{(i)}=0$ denotes its absence. 
Let $\mathrm{D}_U=\big\{ \mathbf{x}_U^{(i)} \big\}_{i=1}^{N_U}$ be an unlabeled dataset of $N_U$ samples, which does not contain labels for its inputs. For  $i$-th unlabeled sample, $p\big(\omega(\mathbf{x}_U^{(i)})\big)=[p_1^{(i)}, \dots, p_C^{(i)}] \in [0, 1]^C$ is the soft prediction (confidence score) of a model, where $\omega(\cdot)$ refers to the weak stochastic data augmentation function. With a chosen score threshold $\tau_c$ for class $c$, the soft prediction $p_c^{(i)}$ is converted to hard pseudo-label by $\Tilde{p}_c^{(i)}=\mathbbm{1}(p_c^{(i)}>\tau_c)$.
SSL task is learning a parameterized model on the combined dataset $\mathrm{D}_L \cup \mathrm{D}_U$.

\subsection{Positive and Negative Pseudo-label Learning}

Selecting suitable unlabeled samples is a critical component in prior SSL methods based on either pseudo-labeling or consistency learning, because predictions on unlabeled samples with high uncertainty will lower performance by confusing model and should be filtered out \cite{zheng2021rectifying, DBLP:conf/iclr/XingAZP20}. In single label classification tasks, a widely adopted solution is selecting unlabeled samples whose maximum class score exceeds a high score threshold, which could be fixed \cite{lee2013pseudo,sohn2020fixmatch} or varying during training \cite{zhang2021flexmatch}. The assumption behind this is that a high score threshold can filter out noisy pseudo labels and leave only high-accuracy ones, leading to a decrease in the confirmation bias \cite{arazo2020pseudo}. Since the probability vector of each sample sums to $1$, a negative pseudo-label of any selected sample must correspond to a probability that is less than $1-\tau$. In practice, with $\tau>0.5$ we technically have a positive score threshold $\tau_+=\tau$ and an implied negative score threshold $\tau_-=1-\tau$. Confidence scores above $\tau_+$ are assigned 1, those below $\tau_-$ are assigned 0, and the interval between $\tau_-$ and $\tau_+$ becomes discarded region. Samples with any score that falls between $\tau_-$ and $\tau_+$ are ignored as a whole.

On the other hand, for multi-label classification, the confidence scores of all classes is no longer required to sum to $1$. As a consequence, we need to explicitly introduce both score thresholds $\tau_\pm$ where $\tau_- < \tau_+$. Let $\mathbf{g}^{(i)}=[g_1^{(i)}, \dots, g_C^{(i)}] \in \{0, 1\}^C$ be the binary vector indicating the pseudo-label selection for the $i$-th unlabeled sample, where $g_c^{(i)}=1$ when $\Tilde{p}_c^{(i)}$ is selected and $g_c^{(i)}=0$ otherwise. This vector is generalized as follows for both single-label and multi-label classification:
\begin{equation}
  g_c^{(i)} = \mathbbm{1}\big[ p_c^{(i)} > \tau_{c,+} \big] + \mathbbm{1}\big[ p_c^{(i)} < \tau_{c,-} \big].
  \label{eq:first_mask}
\end{equation}

Formally, the unsupervised training objective for unlabeled data can be written as
\begin{equation}
  \mathcal{L}_u = \frac{1}{\mu B} \sum_{i=1}^{\mu B} \mathbf{g}^{(i)} \cdot \mathcal{H}\Big(\Tilde{p}\big(\omega(\mathbf{x}_U^{(i)})\big), p\big(\Omega(\mathbf{x}_U^{(i)})\big)\Big),
  \label{eq:loss_u}
\end{equation}
where $B$ is the labeled batch size, $\mu$ is the ratio of unlabeled batch size to labeled batch size, $\mathcal{H}(\cdot, \cdot)$ represents asymmetric loss \cite{ridnik2021asymmetric}, and $\omega(\cdot)$ and $\Omega(\cdot)$ are correspondingly the weak and strong stochastic data augmentation functions.

\subsection{Percentile Thresholds}

As indicated in \cite{zhang2021flexmatch}, the divergent learning difficulties and distribution of different classes make it challenging to choose a universal score threshold for all classes. Instead of the curriculum pseudo-labeling method, here we introduce the concept of percentile thresholds to unify fixed and variable score thresholds.

It is not straightforward to choose an optimal score threshold or interpret the meaning of a chosen one, because the confidence score distribution can be reshaped and shifted qualitatively by changing training components, such as the loss function or labeled data size. To improve the generality and interpretability of confidence-based selection, we set positive and negative percentile thresholds $\kappa_\pm$, and pin down the corresponding score threshold $\tau_\pm$ such that
\begin{equation}
  Prob\Big[p_c(\omega(\mathbf{x}_U^{(i)})) \le \tau_\pm \Big] = \kappa_\pm ~~~~\text{for $\mathbf{x}_U^{(i)}\in \mathrm{D}_U$}.
  \label{eq:tau_pm}
\end{equation}
For each class, if the negative sample ratio in labeled data is higher than the global $\kappa_+$, the local percentile threshold for that class is set to negative sample ratio. In short, we use ground truth ratio in labeled data to avoid have incorrectly high ratio of positive pseudo-labels.

Fig. \ref{fig:thres}(a) depicts two example distribution, which can be of the same class at different time, or of different classes. At early stage of training process, models tend to give conservative predictions, so the resulting score distributions are closer to the blue curve than the red one. On the other hand, if the sample ratio of a class is high, its score distribution would shift right and become more similar to the blue curve. For a given negative percentile threshold $\kappa_-=0.1$, the corresponding score threshold $\tau_-$ are determined so that the area of each green shadow under the distribution curve is $0.1$, as the arrows indicated. Likewise, one can determine the positive score thresholds $\tau_+$ by letting the brown shadow under distribution curves has an area of $1-\kappa_+$.

Now the question becomes how to obtain the score distribution in order to connect percentile thresholds and score thresholds. As the sizes of unlabeled datasets are typically large in SSL scenario, it is unnecessarily expensive to calculate the exact score distribution as that would require running model prediction on the whole unlabeled dataset at every training step. Instead, after dividing the $[0, 1]$ interval into $K$ equal bins, for a given class $c$ and training step $t$, the exponential moving average (EMA) histogram vector $\mathbf{P}_{c(t)}=\Big[P_{c(t)1}, \dots, P_{c(t)K}\Big]$ is used as a coarse-grain estimate of score distribution:
\begin{equation}
\mathbf{P}_{c(t)} = 
\begin{cases}
  [\frac{1}{K}, \dots, \frac{1}{K}], & \text{if } t=0;\\
   \lambda P_{c(t-1)} + (1-\lambda) \frac{1}{\mu B} \mathit{hist}(\{p_{c(t)}^{(i)}\}_{i=1}^{\mu B}), & \text{if } t>0;
\end{cases}
\label{eq:score_dist}
\end{equation}
where $\mathit{hist}$ denotes histogram function with $K$ bins applied on each unlabeled mini-batch, and $\lambda\in[0, 1]$ is a hyperparameter controlling momentum decay. When $\lambda=0$, $\mathbf{P_c}$ is frozen so that score thresholds are fixed at $\tau_\pm=\kappa_\pm$; when $\lambda>0$, the score thresholds are varying automatically over different classes and iterations to make sure only a desired portion of pseudo-labels are selected. The estimate of score distribution does not require any extra inference.

The percentile threshold effectively down-samples for majority classes whose actual positive percentages are higher than $1-\kappa_+$. This feature helps to balance the class distributions in multi-label datasets.

\begin{figure*}
\begin{center}
\includegraphics[width=.97\linewidth]{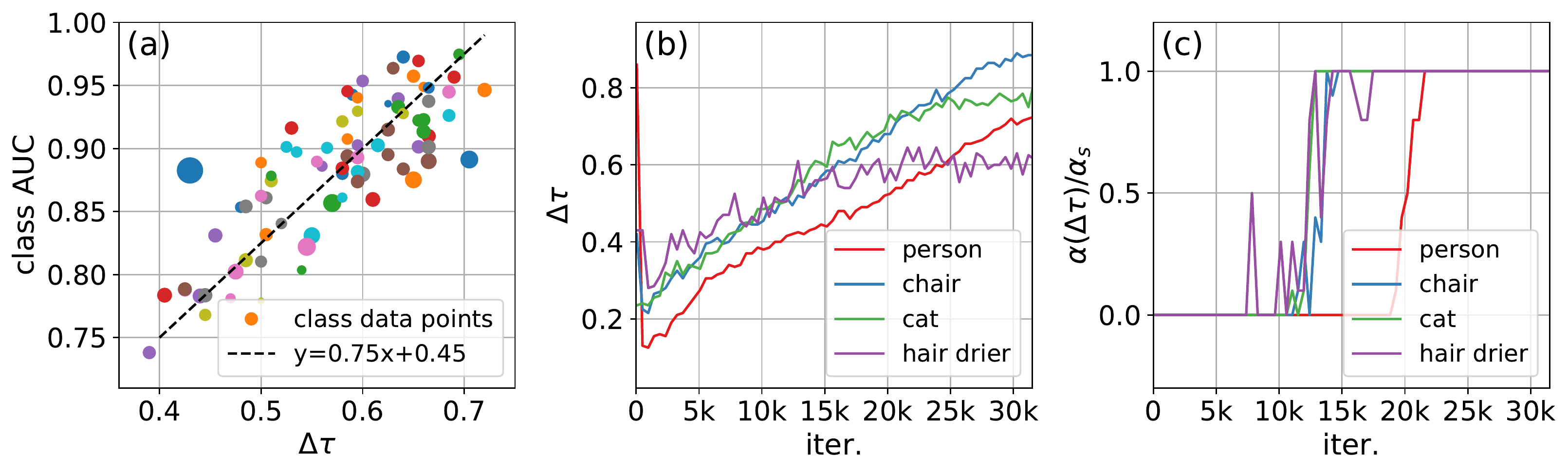}
\end{center}
   \caption{Illustration of dynamic unlabeled loss weights. (a) shows the class AUC of all 80 classes as functions of corresponding $\Delta\tau=\tau_+-\tau_-$ values. The marker sizes are proportional to the fourth roots of class positive sample counts in the unlabeled dataset. The black dashed line is there for visual aid. (b) shows the trace of $\Delta\tau$ from the same experiment as Fig. \ref{fig:thres}, confirmed the trend increasing threshold gap over time. (c) depicts the trace of corresponding unlabeled loss weights in unit of $\alpha_s$ using Eq. \eqref{eq:alpha}, where $\Upsilon_0=0.5$ and $\Upsilon_s=0.55$.}
\label{fig:weight}
\end{figure*}

For illustrative purposes, four classes in MS-COCO datasets are chosen as examples and sorted in descending order of their sample distribution ratio in labeled data as ``person''$>$``chair''$>$``cat''$>$``hair drier'', e.g. half of samples have class ``person'', yet class ``hair drier'' appears in less than $0.2\%$ of samples. In Fig. \ref{fig:thres}(b), we show the trace of positive and negative score thresholds over 32K iterations. The score threshold curves are arranged in the order of class distribution ratio. After a quick warm up of the first $300$ iterations, the positive score thresholds are increasing with some fluctuation, while the negative thresholds are decreasing as training progresses, leading to a gradually wider gap between $\kappa_+$ and $\kappa_-$ over time. To evaluate the learning status of each class, we calculate class ROC-AUC score on COCO val2014 dataset every 1K iterations. For highly imbalanced data, as AUC score is not affected by different sampling rules, it is a better performance measure than accuracy or precision score. Interestingly, as Fig. \ref{fig:thres}(c) shown, the AUC scores of ``person'' and ``cat'' are fairly close to each other after 10K iterations, and higher than that of ``chair'', even when the positive score thresholds have different order. This suggests that the underlying class distribution has a strong effect on the score thresholds, and the number of high-confidence predictions itself is not a good indicator of learning status. We will explore on finding a better indicator in the next subsection.

\subsection{Dynamic Unlabeled Loss Weights}

Prior SSL methods have introduced a time-dependent unlabeled loss weight $\alpha(t)$ to avoid poor local minima \cite{lee2013pseudo, Laine2017TemporalEF}. The intuition of this special treatment for single-label dataset is that even when a score threshold is applied to select only high-score predictions, when a network is poorly calibrated at early stage, it could give a lot of incorrect yet high confidence-score predictions and mislead the optimization process. This treatment becomes even more important for multi-label cases, because when class presence are independent, a model cannot correct the wrong impression of one class from knowledge of other classes. This raises the importance of the model learning status estimate, which can be used to control whether a pseudo-label should be involved in the training. As mentioned previously, the number of high score predictions is not necessarily related to learning status. On the other hand, from Fig. \ref{fig:thres}(b) we notice that while the class ROC-AUC values increase over time, the gaps between positive and negative score thresholds $\Delta\tau=\tau_+-\tau_-$ also increase. Enlightened by this, we then verify that $\Delta\tau$ can be used as a good indicator of learning status by the linear relationship between $\Delta\tau$ and class ROC-AUC scores in Fig. \ref{fig:weight}(a). With $\Delta\tau$ of class $c$ being an indicators of class learning status, we can easily decide its unlabeled loss weight as
\begin{equation}
    \alpha(\Delta\tau)=
    \begin{cases}
    0, & \text{if } \Delta\tau  < \Upsilon_0 \text{ or } t < 300; \\ 
    \alpha_s, & \text{if } \Delta\tau>\Upsilon_s;\\
    \alpha_s \dfrac{\Delta\tau-\Upsilon_0}{\Upsilon_s-\Upsilon_0}, & \text{otherwise};
    \end{cases}
    \label{eq:alpha}
\end{equation}
where $\Upsilon_0$ and $\Upsilon_s$ are, correspondingly, the start and saturate values of threshold gap, and $\alpha_s$ is the saturate unlabeled loss weight. Some example traces of $\alpha(\Delta\tau)$ are shown in Fig. \ref{fig:weight}(c), as each trace behaves similarly to time-controlled unlabeled loss weight, it gains extra flexibility allowing pseudo-labels of different learning difficulties to be involved at different time.


\begin{algorithm*}
\caption{PercentMatch algorithm}\label{alg:pps}
\begin{algorithmic}
\Require a labeled dataset $\mathrm{D}_L$, an unlabeled dataset $\mathrm{D}_U$
\Ensure $0 \le \kappa_- < \kappa_+ \le 1$
\For{c=1 to C}
\State Initialize $\mathbf{P}_{c(0)}$ using Eq. \eqref{eq:score_dist}
\State Initialize $\kappa_{c,+}=max(\kappa_+, \text{ground truth negative ratio in }\mathrm{D}_L)$
\State Initialize $\kappa_{c,-}=min(\kappa_-, \text{ground truth negative ratio in }\mathrm{D}_L)$
\EndFor

\While{not reach the maximum iteration}
    \For{c=1 to C}
    \Comment{Loop over each class}
    \State Convert percentile thresholds $\kappa_{c,\pm}$ to score thresholds $\tau_{c,\pm}$ using Eq. \eqref{eq:tau_pm}
    \State Calculate class unlabeled loss weight $\alpha_c$ using Eq. \eqref{eq:alpha}
    
    \For{i=1 to $\mu B$}
    \Comment{Loop over unlabeled mini-batch}
    \State Obtain predictions $p_c^{(i)}=p_c\big(\omega(\mathbf{x}_U^{(i)})\big)$ and $p_c\big(\Omega(\mathbf{x}_U^{(i)})\big)$
    \State Convert $p_c^{(i)}$ to pseudo-labels $\Tilde{p}_c^{(i)}$ by comparing with $\tau_{c,+}$
    \State Calculate $\mathbf{g}_c^{(i)}$ using Eq. \eqref{eq:first_mask}
    \EndFor
    \State Update $\mathbf{P}_c$ with $\{p_c^{(i)}\}_{i=1}^{\mu B}$ using Eq. \eqref{eq:score_dist}
    \EndFor

    \State Calculate loss using Eqs. \eqref{eq:loss_u}, \eqref{eq:loss_t}, and \eqref{eq:loss_s}
    \State Update model parameters using Adam optimization
\EndWhile
\State \Return model parameters
\end{algorithmic}
\label{alg}
\end{algorithm*}

Finally, the total loss in PercentMatch can be expressed as the weighted combination of supervised and unlabeled loss:
\begin{equation}
\mathcal{L} = \mathcal{L}_s + \alpha \mathcal{L}_u, 
\label{eq:loss_t}
\end{equation}
where $\alpha$ is the ratio of unlabeled loss to supervised loss, and $\mathit{L}_s$ is the supervised loss on labeled data
\begin{equation}
  \mathcal{L}_s = \frac{1}{B} \sum_{i=1}^{ B} \mathcal{H}\Big(\mathbf{y}^{(i)}, p(\omega(\mathbf{x}^{(i)}))\Big),
  \label{eq:loss_s}
\end{equation}

The pseudo-code of PercentMatch algorithm is provided in Alg. \ref{alg}. At each training step, PercentMatch adds a negligibly small computational cost compared to FixMatch, mainly due to updating the score histogram vectors for unlabeled mini-batch and calculating dynamic unlabeled loss weights.



\section{Experiments}

\begin{table*}
  \centering
  \begin{tabular}{c | c | c | c | c | c | c}
    \hline
    \multirow{2}{*}{Method} 
    & \multirow{2}{*}{Backbone} 
    & \multirow{2}{*}{Size}
    & \multicolumn{2}{c}{10\% labeled} 
    & \multicolumn{2}{c}{20\% labeled}\\
     & & & mAP & AUC
    & mAP & AUC\\
    \hline
    \bottomrule
    
    Supervised* &  ResNet50 & 224 & 18.36 & - & 28.84 & - \\
    PL*\cite{lee2013pseudo}& ResNet50 & 224 & 27.44 & - & 34.84 & -\\
    MixMatch*\cite{berthelot2019mixmatch} & ResNet50 & 224 & 29.57 & - & 37.02 & -\\
    MT*\cite{t2017mean} & ResNet50 & 224 & 32.55 & - & 39.62 & -\\
    UPS*\cite{rizve2021in} & ResNet50 & 224 & \textbf{34.22} & - & \underline{40.34} & -\\
    \hline
    PercentMatch & ResNet50 & 224 & \underline{33.43} & 81.12 & \textbf{42.38} & 85.69 \\
    PercentMatch & ResNet50 & 448 & 37.25 & 83.26 & 45.19 & 87.14\\
    \bottomrule
  \end{tabular}
  \caption{mAP and AUC scores (in \%) on the Pascal VOC2007 test set. Methods with * use scores reported in \cite{rizve2021in}. All models used start with randomly initialized weights and no pre-training. Bold font indicates the best mAP score of image size 224, and underline indicates the second best.}
  \label{tab:pascal}
\end{table*}

\subsection{Implementation Details}

We evaluate our method on Pascal VOC2007 and MS-COCO 2014 datasets with various labeled data ratio. We use ResNet50 as model backbone and the same hyperparameters for all experiments across both datasets unless specifically indicated. All model weights are initialized randomly without pre-training for a fair comparison.

\textbf{Pascal VOC2007} is a widely used multi-label classification dataset. It has 5,011 train-val images and 4,952 test images, each of which can contain 1 to 6 classes out of 20 classes. The imbalance ratio, which is the ratio of the sample size of the most majority class and that of the most minority class as a measure of class-imbalance extent. For example, the imbalance ratio of Pascal VOC2007 train-val set is $2008/96=20.9$.

\begin{table*}
  \centering
  \begin{tabular}{c | c | c | c | c | c | c | c | c}
    \hline
    \multirow{2}{*}{Method} 
    & \multirow{2}{*}{Backbone} 
    & \multirow{2}{*}{Size}
    & \multicolumn{2}{c}{2\% labeled}
    & \multicolumn{2}{c}{5\% labeled}
    & \multicolumn{2}{c}{10\% labeled}\\
     & & & mAP & AUC
    & mAP & AUC & mAP & AUC\\
    \hline
    \bottomrule
    
    FixMatch & ResNet50 & 224 & 21.05 & 81.54 & 29.33 & 86.97 & 35.26 & 89.71\\
    PercentMatch & ResNet50 & 224 & 21.54 & 80.26 & 30.59 & 86.81 & 37.64 & 89.82 \\
    PercentMatch & ResNet50 & 448 & 23.87 & 81.24 & 32.29 & 86.96 & 41.33 & 90.41\\
    \bottomrule
  \end{tabular}
  \caption{mAP and AUC scores (in \%) on the MS-COCO val2014 set. All models used are initialized randomly.}
  \label{tab:example}
\end{table*}

We use ResNet50 as the backbone with ML-decoder head \cite{ridnik2021ml}, and train our model using Adam optimizer and one-cycle policy for 7K iterations, with maximal learning rate of $3\text{e-}4$. For regularization, we apply the common ImageNet statistics normalization to all input images, along with the data augmentation used in contrastive learning \cite{chen2020simple}, which consists of center crop, random horizontal flip, and random cutout. Besides, we add three and eight random augmentation operations to the weak and strong augmentation for unlabeled data, correspondingly. The labeled batch size $B$ is $36$, and $\mu$ is set to $1$. The saturate unlabeled loss weight $\alpha_s=1$, and the start and saturate threshold gaps are $\Upsilon_0=0.5$ and $\Upsilon_s=0.55$. The percentile thresholds are $\kappa_-=0.1$, $\kappa_+=0.98$. Our experiments show that the final performance is not very sensitive to these hyperparameters.

\textbf{MS-COCO 2014} contains 82,783 training images and 40,504 validation images. There are 80 different labels which can appear at the same image, forming a multi-label classification problem. The training images can have 0 to 18 different classes. The most predominant class, person, appears in 45,174, and the most rare class, hair drier, only appears 128 times, results in an imbalance ratio of $353$. All other hyperparameters are the same as experiments on Pascal VOC2007, except that we use 32K iterations for MS-COCO in order to accommodate different dataset size. 

\subsection{Results}

For Pascal VOC2007 dataset, we implement two experiments with $10\%$ (500 samples) and $20\%$ (1000 samples) randomly selected from the train-val split (5011 samples) as the labeled set, correspondingly, and the rest as the unlabeled set.  To the best of our knowledge, the result on Pascal VOC2007 reported in \cite{rizve2021in} is the only benchmark on \textit{multi-label semi-supervised classification}, so we use the same model backbone and input size for fair comparison. The results are reported in Table \ref{tab:pascal}. Our method leads to a $2.04\%$ mAP score improvement for Pascal VOC2007 $20\%$ labeled data, and a close score on $10\%$ labeled data, compared with the previous SOTA results. It is worth mentioning that UPS is much slower than FixMatch and our method, as it performs a big number of inference on each sample for uncertainty calculation and the training process need to be repeated for multiple generations of teacher-student cycles.

For MS-COCO dataset, we implement three experiments with $2\%$ (1640 samples), $5\%$ (4100 samples) and $10\%$ (8200 samples) randomly selected from the training images as the labeled set, correspondingly, and the rest as the unlabeled set. We test the model performance on the whole validation set. As we cannot find any reported MS-COCO benchmark on \textit{multi-label semi-supervised classification}, we re-implement FixMatch using the same hyperparameters, except percentile thresholds $\kappa_\pm$ are replaced with fixed confidence thresholds $\tau_+=0.95$ and $\tau_-=0$ following the original paper. The proposed PercentMatch can consistently outperform FixMatch on the mAP score, confirming that the dynamic thresholding is important to the overall performance. We notice that PercentMatch gains bigger improvement when the
labeled ratio is big comparing to FixMatch, partially due to the fact that higher labeled ratio leads to earlier involvement of unlabeled loss by exceeding the uniform $\Upsilon_0$ we use for all experiments.

\section{Conclusion}

While previous study has developed different SSL methodologies to leverage unlabeled data, most of them focus on single-label classification. In this study, we analyze some special challenges in multi-label semi-supervised classification with illustrative examples. To address these challenges, we propose PercentMatch, an percentile-based dynamic thresholding framework that maintains the simplicity of FixMatch and naturally introduces dynamical score thresholds for positive and negative pseudo-labels. In addition, the adaptive unlabeled loss weights decided by the difference of two score thresholds helps to reduce noise introduced by incorrect pseudo-labels at early training stage. Our method leads to strong performance and fast convergence on standard multi-label datasets. We believe percentile-based thresholding has more potential usages in SSL, for example by gradually lower the positive percentile threshold one can introduce curriculum learning to the framework. We hope our work could inspire more future work in the related field.

{\small
\bibliographystyle{ieee_fullname}
\bibliography{egbib}
}

\end{document}